\documentclass[10pt,letterpaper]{article}

\usepackage{cogsci}
\usepackage{graphicx}
\usepackage{subcaption}
\usepackage{bm}

\usepackage{fancyhdr}
\fancypagestyle{title}{%
  \setlength{\headheight}{28pt}%
  \fancyhf{}
  \fancyhead[C]{\small Proceedings of the 42nd Annual Meeting of the Cognitive Science Society (2021). \\}
}%

\cogscifinalcopy 

\usepackage{pslatex}
\usepackage{apacite}
\usepackage{float} 

\title{Modelling the development of counting with memory-augmented neural networks}
 
\author{{\large \bf Zack Dulberg (zdulberg@princeton.edu)} \\
  Princeton Neuroscience Institute, Princeton, NJ 
  \AND {\large \bf Taylor Webb (taylor.w.webb@gmail.com)} \\
  University of California Los Angeles, Los Angeles, CA
  \AND {\large \bf Jonathan Cohen (jdc@princeton.edu)} \\
  Princeton Neuroscience Institute, Princeton, NJ }

\begin{document}

\maketitle
\thispagestyle{title}

\begin{abstract}
Learning to count is an important example of the broader human capacity for systematic generalization, and the development of counting is often characterized by an inflection point when children rapidly acquire proficiency with the procedures that support this ability. We aimed to model this process by training a reinforcement learning agent to select N items from a binary vector when instructed (known as the give-$N$ task). We found that a memory-augmented modular network architecture based on the recently proposed Emergent Symbol Binding Network (ESBN) exhibited an inflection during learning that resembled human development. This model was also capable of systematic extrapolation outside the range of its training set - for example, trained only to select between 1 and 10 items, it could succeed at selecting 11 to 15 items as long as it could make use of an arbitrary count sequence of at least that length. The close parallels to child development and the capacity for extrapolation suggest that our model could shed light on the emergence of systematicity in humans.

\end{abstract}

\section{Introduction}

Humans are capable of systematic generalization, that is, performing well outside the range of values on which they were trained \cite{marcus2001algebraic, chollet2019measure}. For example, a human could fetch 12 apples if asked, despite having only ever grabbed up to 9 apples in the past. Although this capacity falls short of perfect systematicity \cite{lake2019human}, artificial neural networks have much greater difficulty performing well in contexts outside the convexity of their training data \cite{lake2018generalization,barrett2018measuring}. Learning to count is one of the earliest systematic behaviours acquired in human development, and is foundational with respect to further development of abstract procedures like mathematics. Here, we present a counting model that exhibits both a developmental trajectory similar to humans as well as systematicity. 

The development of counting in childhood has previously been summarized by the knower-level theory \cite{wynn1990children,wynn1992children,carey2001cognitive,sarnecka2006development}. This theory suggests a number of distinct stages in acquiring an understanding of the cardinal meaning of numbers. A child begins as a pre-numeral knower, with no understanding of cardinality, and then becomes a subset-knower, learning subsequent numbers in order (i.e., becomes a one-knower, then a two-knower, then a three-knower, etc). Around the time a child becomes a five-knower, there appears to be an inductive transition, after which the child becomes a cardinal-principle-knower (CP-knower), and understands the cardinal meaning of numbers as high as they can count. Some data have supported this view, though children pass through knower-levels at different rates \cite{sarnecka2009levels}, and early stages may be more approximate than previously thought \cite{wagner2019children}, while other data have called into question whether a true semantic induction underlies this apparent transition \cite{davidson2012does}.

The knower-level theory has generally been based on data from the widely used give-$N$ task \cite{wynn1990children,wynn1992children,frye1989young}. In this task, a child is instructed to 'give $N$ objects', and is typically considered an $N$-knower if they can select the correct number of objects twice as often as they make an error (66 \% accuracy). Our goal was to train a neural network to perform the give-$N$ task and examine its performance and developmental trajectory. We suggest the give-$N$ task is most realistically modelled using a reinforcement learning framework. First, the task can be successfully completed in many different ways (i.e. by selecting objects in various orders), so a reinforcement rather than a more specific instructive signal is appropriate. Furthermore, children receive a substantial amount of reinforcement from adults when learning tasks like these ("good job!"). Finally, the give-$N$ task can easily be expressed as a decision process with discrete actions, which therefore lends itself naturally to reinforcement learning.

We report simulations of a model that, trained to perform the give-$N$ task using reinforcement learning, displays both an inflection point consistent with the human developmental trajectory, as well as the capacity for systematic extrapolation when presented with stimuli out of the range of its training. The model extended the Emergent Symbol Binding Network (ESBN) \cite{webb2021emergent}, in which internal representations of a control network were separated from task inputs, interacting only through binding in a differentiable external memory module. We developed a variant of the ESBN within a reinforcement learning framework to address how its capacity for symbol-like behavior might support the development of systematic counting proficiency. We also softened the separation constraint, giving the ESBN access to both internal and input streams (and thus the option to over-fit to training inputs), and it still exhibited symbol-like behavior and the capacity for systematic generalization.  Baseline models did not display inflection or extrapolation, suggesting variable binding and dot-product similarity evaluation as the relevant architectural inductive biases that promoted systematicity and emulated human development. This work also supports the broader idea that the development of abstract conceptual knowledge may be supported by the learning of systematic procedures.

\section{Related Work}
There have been various attempts to model counting. One line of inquiry investigated the ability of neural networks to estimate numerosity by, for example, looking at an image of objects \cite{chen2018can,stoianov2012emergence,zorzi2018emergentist}. Another study used counting equivariance relations to support the learning of visual representations \cite{noroozi2017representation}. However, such perceptual tasks do not address the capacity for systematic counting behavior. Other work has shown that recurrent neural networks can keep track of counts in order to predict the next character in a string derived from some grammar \cite{rodriguez1999recurrent,rodriguez2001simple}. Though accuracy was limited, some of these networks could extrapolate to longer strings than those on which they were trained. However, this was a character prediction task rather than a counting task, making the parallel to human counting development unclear. Finally, Bayesian models have been proposed to address the development of systematic counting in humans \cite{piantadosi2012bootstrapping,lee2010model}. These models generally assumed knower-levels (including CP-knower) as primitive hypotheses. The model we propose does not include such primitives, seeking to explain systematicity as an emergent property that arises through learning.


Some neural network modeling studies have directly addressed learning to count. \citeA{luteaching} introduced a feed-forward network with a visual attention mechanism that was trained through a combination of reinforcement learning and ‘social scaffolding’ (demonstrations from a teacher) to perform the how-many task (a simpler counting task that children often master before they can perform the give-$N$ task). \citeA{fang2018can} developed an extension of this model that used a recurrent network and operated over more realistic two-dimensional inputs. Although this network learned numbers in the correct order (passing appropriately through subset-knower levels), it did not display an inflection point for higher numbers, and was not tested on extrapolation for numbers outside the range on which it was trained. \citeA{sabathiel2020computational} expanded further on this work by training a recurrent convolutional network to perform a variety of counting tasks, including give-$N$. However, the give-$N$ task did not display a familiar developmental trajectory (for example, it learned to give-1 last rather than first), and this model was also not tested on numbers outside the training range. A common element of these studies was the use of teacher-guided learning; that is, networks were trained to imitate a specific way of solving the task, rather than learning from reinforcement alone. Imitation has in fact been shown to improve learning of multiple tasks in a rich 3D environment, but interestingly, in that work, agents still struggled most with learning to count~\cite{abramson2020imitating}. Although imitation likely plays a role in the development of many skills, here we investigated to what extent a model trained from reinforcement alone could account for the relevant developmental phenomena.

\section{Methods}
\subsection{Environment}

We represented a set of objects in the give-$N$ task as a binary vector in which 1's corresponded to the presence of an object at a given location. For example, the vector [0,1,0,1,1,0] indicated that there were objects at locations 2, 4, and 5. Vectors were of length 40 so that the space of possible object arrangements was combinatorically large ($2^{40}$ states), in order to prevent memorization. All models were trained on the give-$N$ task as follows: given an initial object vector $\bm{o_0}$, and an instruction to give $N$ objects (represented by the one-hot vector $\bm{x_N}$), the model should select $N$ unique objects one at a time from the object vector to produce a correct response. If a location with an object was selected at time $t -1$, the value at that location changed from 1 to 0 in $\bm{o_t}$. The binary object vector can be interpreted as the output of an object segmentation model; focusing on this intermediate level of abstraction was motivated by the desire to understand the acquisition of counting competency separately from the details of sensory processing and object segmentation.

\subsection{Model}

\begin{figure}[h]
\centering
\begin{subfigure}[t]{0.39\linewidth}
  \centering
  \includegraphics[width=\linewidth]{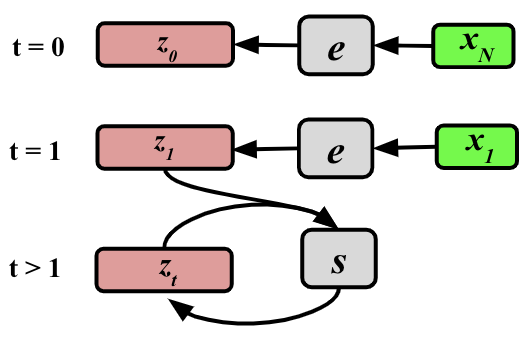}  
  \subcaption{Counter}
  \label{timescheme}
\end{subfigure}%
\begin{subfigure}[t]{0.60\linewidth}
  \centering
  \includegraphics[width=\linewidth]{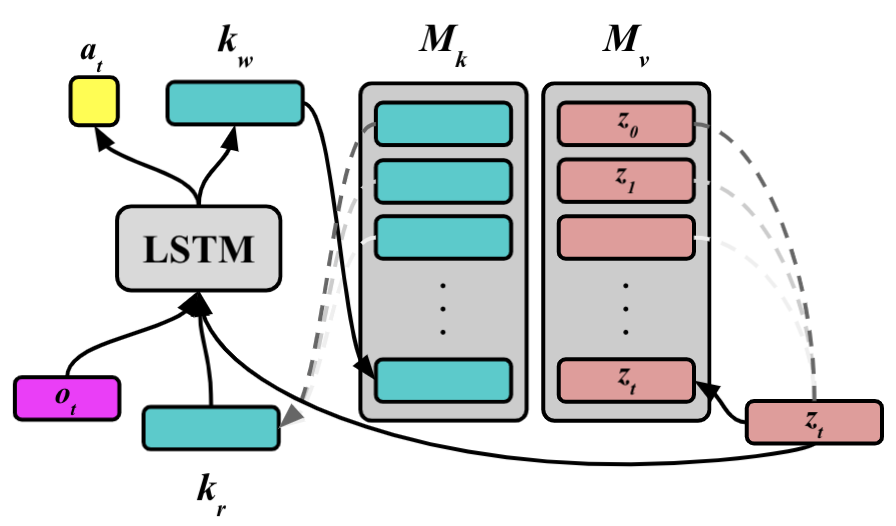}  
  \subcaption{ESBN}
  \label{mscheme}
\end{subfigure}
\begin{subfigure}[t]{0.3\linewidth}
  \centering
  \includegraphics[width=\linewidth]{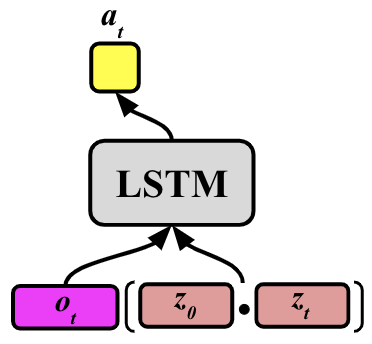}
  \subcaption{\footnotesize Dot-product}
  \label{dscheme}
\end{subfigure}
\begin{subfigure}[t]{0.23\linewidth}
  \centering
  \includegraphics[width=\linewidth]{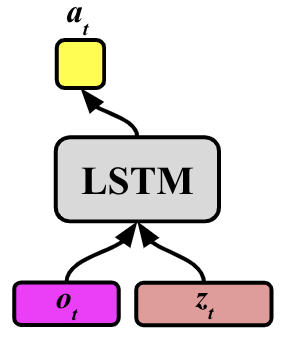}
  \subcaption{\small LSTM}
  \label{bscheme}
\end{subfigure}
\begin{subfigure}[t]{0.42\linewidth}
  \centering
  \includegraphics[width=\linewidth]{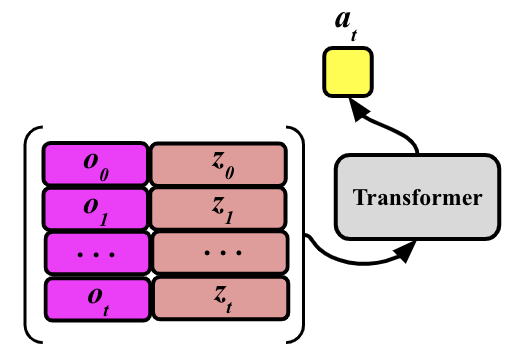}
  \subcaption{\small Transformer}
  \label{tscheme}
\end{subfigure}
\caption{Schematics. Temporal count sequence shown in (a), spatial architecture of models in (b-e). The counter (a) consisted of the encoder $e$ (which translated one-hot vectors $\bm{x_N}$ / $\bm{x_1}$ into count embeddings $\bm{z_0}$ / $\bm{z_1}$), and the successor function $s$ (which iteratively produced $\bm{z_t}$ from $\bm{z_{t-1}}$ starting with $\bm{z_1}$). The ESBN model (b) consisted of an LSTM controller which at each time step received a concatenation of key $\bm{k_{r_t}}$ (read from memory), object vector $\bm{o_t}$ and count embedding $\bm{z_t}$ as inputs, wrote key $\bm{k_{w_t}}$ to memory, and selected action $\bm{a_t}$. The count embeddings also interacted with the controller via a key/value memory system indicated by matrices $\bm{M_k}$/$\bm{M_v}$. In the dot product model (c), only the cosine similarity (dot symbol in brackets) of each count embedding $\bm{z_t}$ with the instruction embedding $\bm{z_0}$ was input to the LSTM controller along with $\bm{o_t}$. In the LSTM baseline (d), $\bm{z_t}$ and $\bm{o_t}$ were input directly into the LSTM controller.  In the transformer baseline (e), the full history of count embeddings and object vectors up until time $t$ was input into a transformer layer.}
\label{scheme}
\end{figure}

First, we pre-trained a counter module that consisted of an encoding function $e$ (a 128 unit linear projection), a successor function $s$ (another 128 unit linear projection), and a decoding function $d$ (a 15 unit linear projection). Given a 15-unit one-hot vector $\bm{x_n}$ representing an integer $n \in\{1,15\}$, the counter was trained to produce an embedding $\bm{z}$ such that $\bm{z} = e(\bm{x_n})$, $\bm{x_n} = d(e(\bm{x_n}))$, and $\bm{x_{n+1}} = d(s(e(\bm{x_n})))$. In this way, $s$ was a successor operation that could iterate through a learned sequence of $\bm{z}$ embeddings, $e$ was an encoder that could translate a one-hot instruction into one of the embeddings in that sequence, and $d$ was a decoder that could translate embeddings back into one-hots (in the our models, $d$ only played a role during pre-training).  These components implemented a counting capability assumed to have been acquired by children at the time they are asked to perform the give-$N$ task. 

The temporal structure of the counter outputs was the same for all models (Fig. \ref{timescheme}), where $\bm{z_t}$ represented the embedding produced by the counter and passed to the networks at teach time step $t$. At $t = 0$,  the task instruction $\bm{x_N}$ was passed through the encoder $e$ to generate $\bm{z_0}$. At $t = 1$, the one-hot representing the start of the count sequence, $\bm{x_1}$, was passed through $e$ to generate $\bm{z_1}$. For each subsequent time step, the embedding from the previous time step $\bm{z_{t-1}}$ was passed through the successor function $s$ to generate the embedding for the next number in the count list $\bm{z_t} = s(\bm{z_{t-1}})$ (representing the counter reciting its learned count sequence). 

The ESBN model consisted of a set of components outlined in Fig. \ref{mscheme}. The controller was an LSTM \cite{hochreiter1997long} augmented with a differentiable (external) memory separated into keys ($\bm{M_k}$) and values ($\bm{M_v}$). The memory was initialized with one learned key/value pair so it did not start out empty, and an additional key/value pair ($\bm{k_{w_t}}$ / $\bm{z_t}$) was written to memory at each subsequent time-step. Note the value here was the $\bm{z_t}$ just described.

As in the original ESBN, the LSTM controller had a single layer with 512 units. The controller received a concatenation of three vectors as input at each time step: $\bm{k_{r_t}}$, $\bm{z_t}$, and $\bm{o_t}$. The input $\bm{k_{r_t}}$ was a key initialized to zeros at $t = 0$, and retrieved from memory at each subsequent time-step. To retrieve $\bm{k_{r_t}}$, the cosine similarity was calculated between $\bm{z_t}$ and all previous values in memory $\bm{M_{v_{t-1}}}$, passed through a softmax to produce a set of weights $\bm{w_{k_t}}$, and used to calculate a weighted sum over $\bm{M_{k_{t-1}}}$, the keys in memory. The second input to the controller, $\bm{z_t}$, was the count embedding itself (so that the decision to use $\bm{k_{r_t}}$, $\bm{z_t}$ or some combination thereof to solve the task was left up to the model). The third input was the object vector $\bm{o_t}$. 


The ESBN model had two output heads. The action head was a fully-connected layer with input size $512$ and output size 41 (corresponding to the 40 possible object locations, plus an additional \textit{done} action). A softmax activation was applied to the output logits to produce a vector of action probabilities. The controller also had a key output head (256 units with ReLU nonlinearities) that produced the key $\bm{k_{w_t}}$ written to memory at each time step .


We compared the ESBN model to a set of models that lacked key/value memory modules.  The dot-product model (Fig. \ref{dscheme}) was meant to elucidate the role of similarity-based memory retrieval in the success of the ESBN model, albeit in a manner that is specifically tailored to the counting task (unlike the original ESBN). We computed the cosine-similarity between $\bm{z_t}$ and $\bm{z_0}$ directly, so the controller only received $\bm{o_t}$ and the scalar similarity score as inputs. In the LSTM baseline (Fig. \ref{bscheme}), the controller simply received $\bm{o_t}$ and $\bm{z_t}$ as inputs.  Finally, in the transformer baseline, the controller was a single transformer layer \cite{vaswani2017attention} (8 self-attention heads, 512-unit MLP, positional encoding) which at each time step received the entire past sequence of $\bm{o_{0..t}}$ and $\bm{z_{0..t}}$, concatenated as shown in Fig. \ref{tscheme}.

\subsection{Training}

\subsubsection{Pre-training Counter}

The counter was pre-trained on its auto-encoding and successor functions. Given a one-hot input vector $\bm{x_n}$ where $n \sim $ Uniform$(1,15)$, the counter produced an output of $d(s^i(e(\bm{x_n})))$. $s^i$ represented iterating $s$, the successor function, $i$ times in a row, with $i \sim $ Uniform$(0, 15-n)$. This allowed for interleaved learning of both encoding and decoding between one-hot space and embedding space, as well as iterating through the learned embeddings sequentially with the successor function. The loss was computed as the mean-squared-error between the output vector and the desired one-hot vector $\bm{x_{n+i}}$.  As well, a similarity penalty on the embeddings was added to the loss (for $i \neq 0$) as the dot product $e(\bm{x_n}) \cdot s^i(e(\bm{x_n}))$. Without this penalty, repeated applications of the successor function caused embeddings to drift apart from their corresponding one-hot encodings. The Adam optimizer \cite{kingma2014adam} was used to perform weight updates on mini-batches of 1 with a learning rate of $10^{-4}$. The weights of the counter network were frozen before being used in the subsequent reinforcement learning task.

\subsubsection{Reinforcement Learning of give-$N$ Task}

During training, action $a_t$ was sampled at each time-step from a categorical distribution using action probabilities produced by our models. We used a two-step training curriculum. 

In the first step, agents were trained only to select 1s and not 0s. The object vector was the only input (all other input units were set to zero), and the agent received a reward of 0 if it selected an object slot that contained a 1 and a reward of -1 otherwise. Episodes ended after 20 time-steps.

In the second step, we switched to the give-$N$ task. Each training episode started by randomly selecting an integer $N$ between 1 and $N_{max}$, represented as the one-hot instruction vector $\bm{x_N}$. $N_{max}$ was initialized to 1, and incremented by 1 once the network achieved at least $66\%$ accuracy on give-$N_{max}$. This curriculum progressed until $N_{max}$ was fixed at 10. Having selected $N$ for a given episode, the object vector was populated with $j$ objects, where $j$ $\sim$ Uniform ($N+10, 35$). This was done to reduce the correlation between $N$ and the number of objects in the object vector, while keeping the space of possible object arrangements very large.

 If the agent selected an object location containing a $1$, that object was replaced with a $0$ on the next time step, and the agent received a reward of $0$. If the agent selected a location containing a $0$, it received a reward of $-1$. Finally, if the agent selected \textit{done} (ending the episode), it received a reward of $+5$ if it had by that time selected exactly $N$ objects, and otherwise a reward of $- | N - n|$ if it had selected $n \neq N$ objects. 

At the end of each episode, one gradient descent step in weight space was performed according to the REINFORCE policy gradient algorithm \cite{williams1992simple} (chosen because it was the simplest algorithm capable of learning the task). The Adam optimizer with a learning rate of $5\times10^{-5}$ was used for a total of 500,000 episodes (50,000 episodes on step 1 and the remaining 450,000 episodes on give-$N$). A set of 30 randomly initialized models were trained for each condition.

\subsection{Testing}
In order to track developmental trajectories, models were tested on give-$N$ at check-points during training every 1000 episodes. At each check-point, an accuracy score was produced by calculating the proportion of correct responses out of 30 new object vectors (unseen during training) for each requested $N \in \{1,10\}$. Actions were selected during testing based on the maximum action probability rather than categorical sampling. A correct response was defined as receiving no negative rewards during the complete episode (based on our reward scheme, this meant the agent selected exactly $N$ objects, and did not select any empty locations, before indicating \textit{done}). The episode at which training accuracy exceeded a threshold of $66\%$ was recorded for each $N \in \{1,10\}$. This produced a developmental trajectory for the accuracy of each model, and we compared how well these trajectories were fit by linear, exponential, logarithmic or sigmoidal functions using the Bayesian information criterion \cite{schwarz1978estimating}. 

\begin{figure}[h]
\centering
\begin{subfigure}[t]{0.49\linewidth}
  \centering
  \includegraphics[width=\linewidth]{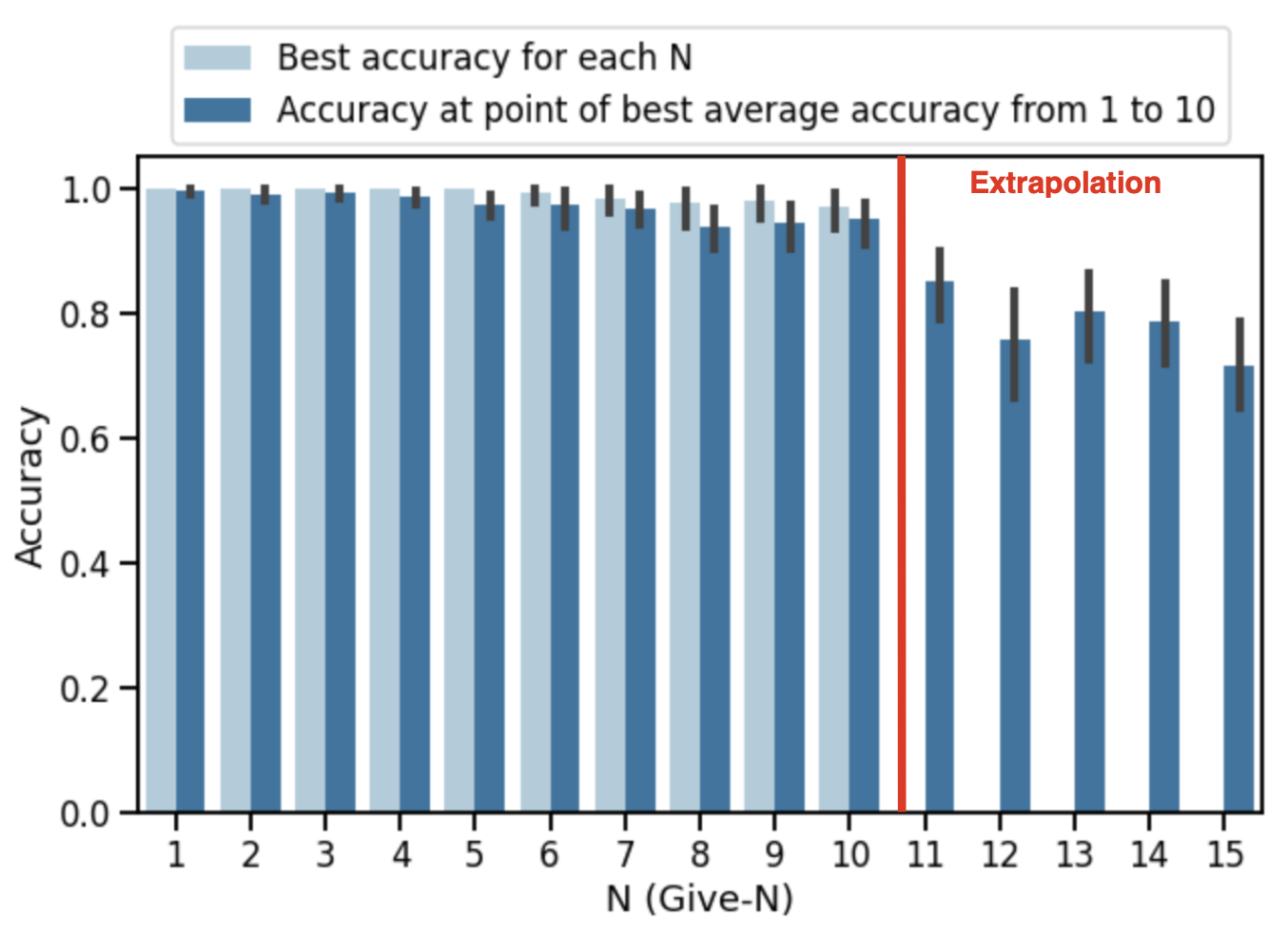}  
  \subcaption{ESBN Model}
  \label{mextrap}
\end{subfigure}%
\begin{subfigure}[t]{0.49\linewidth}
  \centering
  \includegraphics[width=\linewidth]{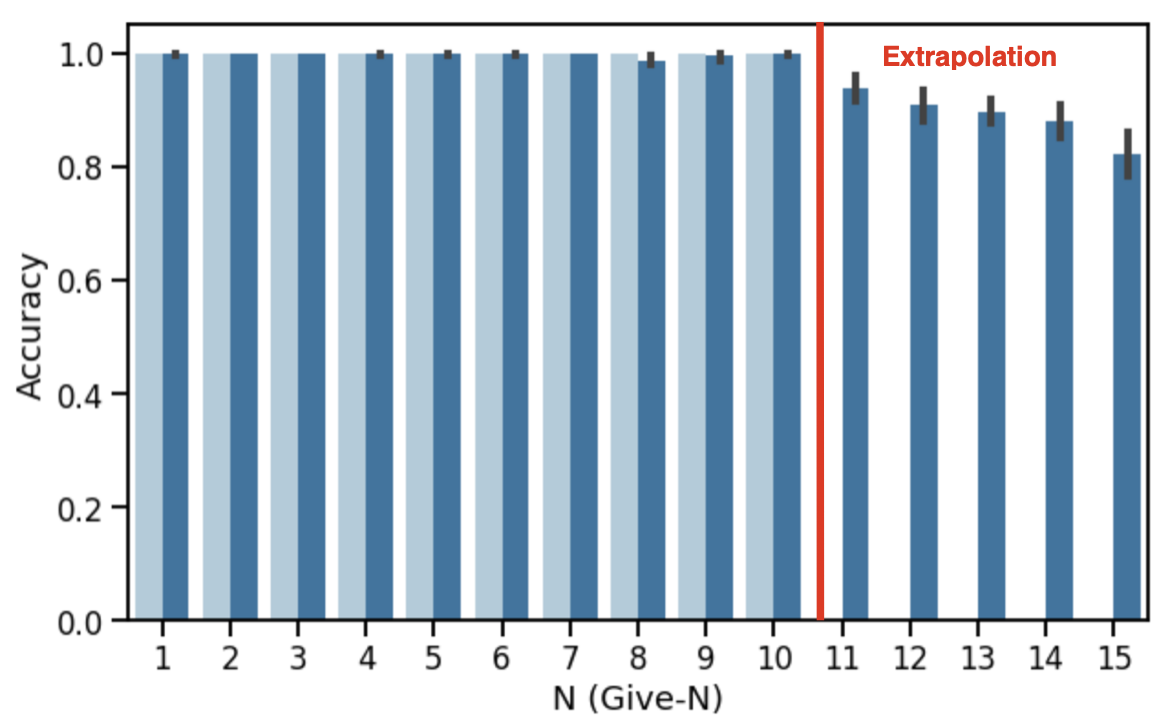}
  \subcaption{Dot-product Model}
  \label{dextrap}
\end{subfigure}
\begin{subfigure}[t]{0.49\linewidth}
  \centering
  \includegraphics[width=\linewidth]{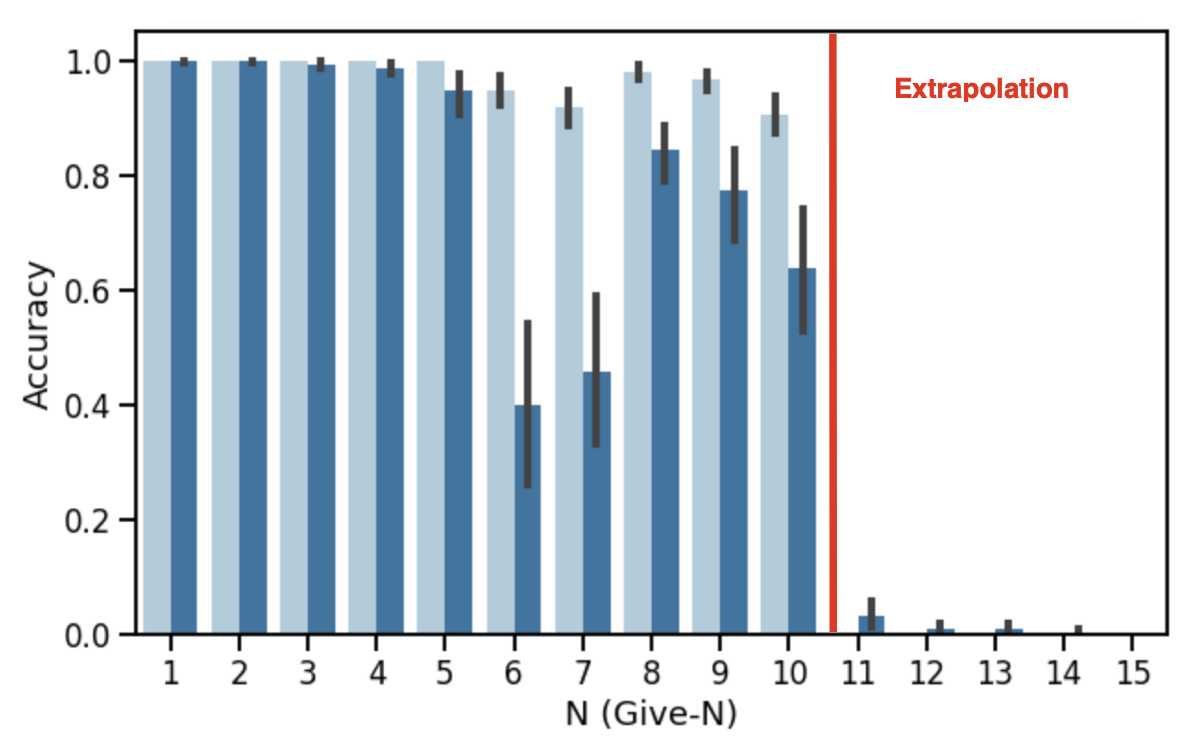}
  \subcaption{LSTM Baseline}
  \label{bextrap}
\end{subfigure}%
\begin{subfigure}[t]{0.49\linewidth}
  \centering
  \includegraphics[width=\linewidth]{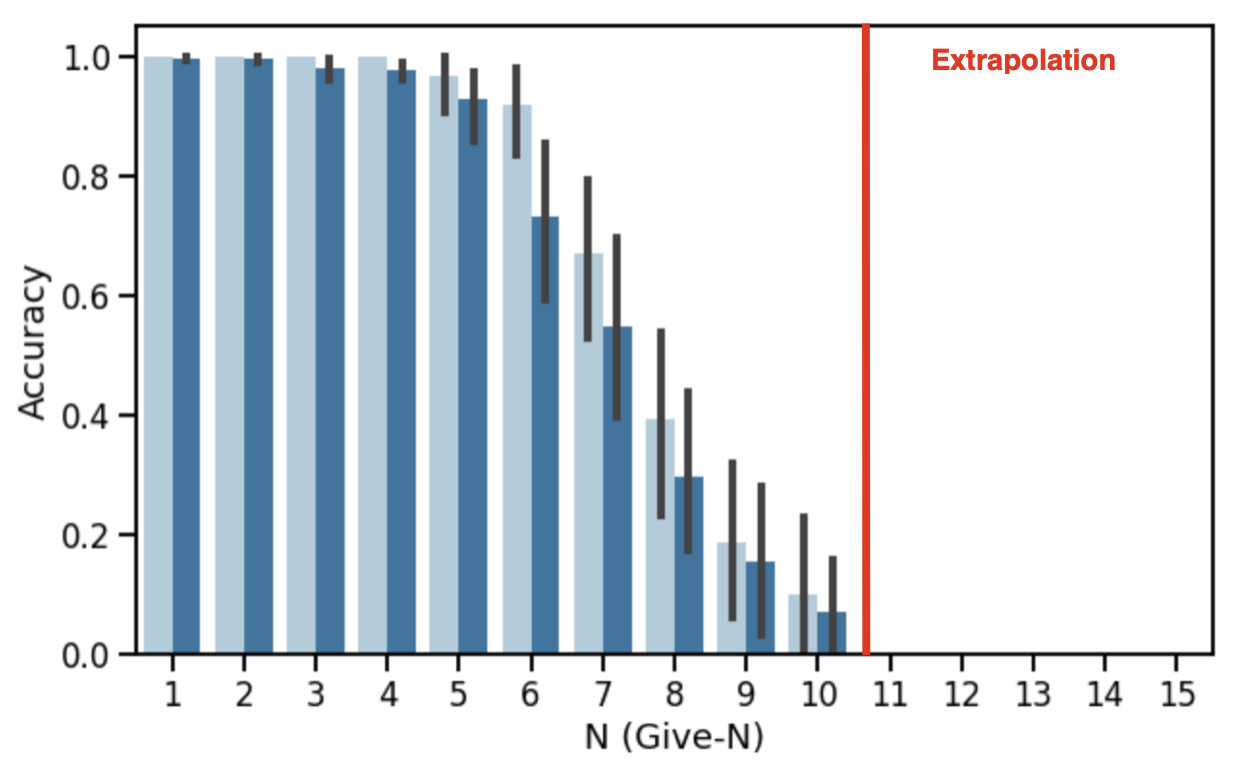}
  \subcaption{Transformer Baseline}
  \label{textrap}
\end{subfigure}
\caption{Accuracy on the give-$N$ task for all trained models. Two bars are displayed for each $N$: the point during training with the highest average accuracy across N from 1 to 10 (dark blue) and the best accuracy for each N at any point in training (light blue, left out for extrapolation to avoid using extrapolation performance to select when to test the model). Bars in each subplot represent mean and standard error across an ensemble of 30 trained models. The red line separates the training regime (left) from the extrapolation regime (right).}
\label{extrap}
\end{figure}

Finally, best performance as well as extrapolation performance was determined by calculating the accuracy of each model at the episode during training that had the highest average accuracy across $N \in \{1,10\}$. The extrapolation set was defined as $N = \{11,15\}$, a set of instructions for give-$N$ that was never presented to the model during training, but was nevertheless tested for accuracy at each checkpoint. Occasionally, our models exhibited unstable behavior (i.e., initially learning the task well, but then dropping in performance prior to the final episode).  This required us to select an appropriate testing point, intentionally not using accuracy on the extrapolation set to determine this point in order to avoid test-set leakage into our results. In order to evaluate more modest success, particularly for the baseline models, we also report the best performance for each $N \in \{1,10\}$ at any point during training individually.

\begin{figure}[h]
\centering
\begin{subfigure}{0.49\linewidth}
  \centering
  \includegraphics[width=\linewidth]{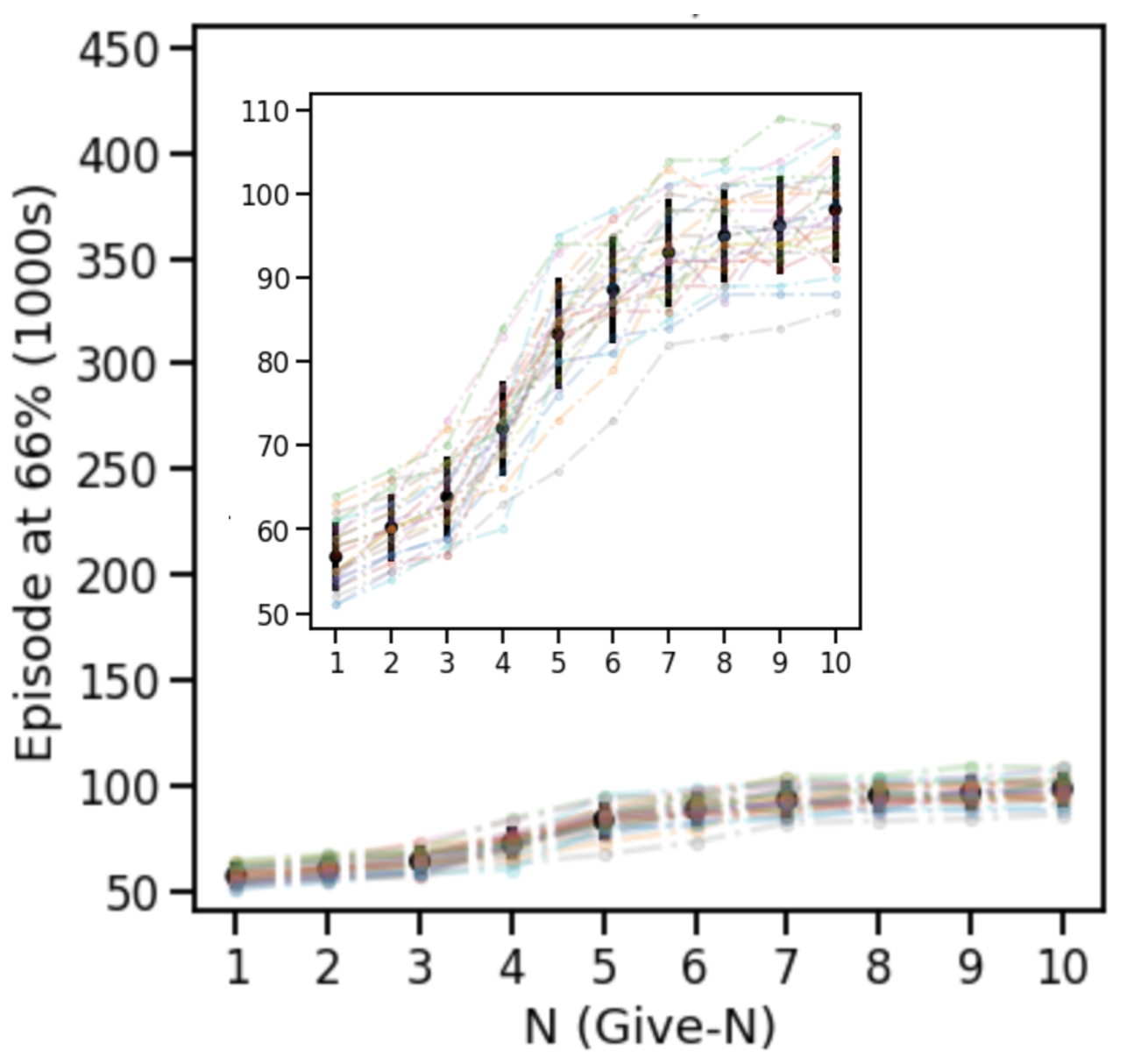}  
  \subcaption{ESBN Model}
  \label{mtraj}
\end{subfigure}%
\begin{subfigure}{0.49\linewidth}
  \centering
  \includegraphics[width=\linewidth]{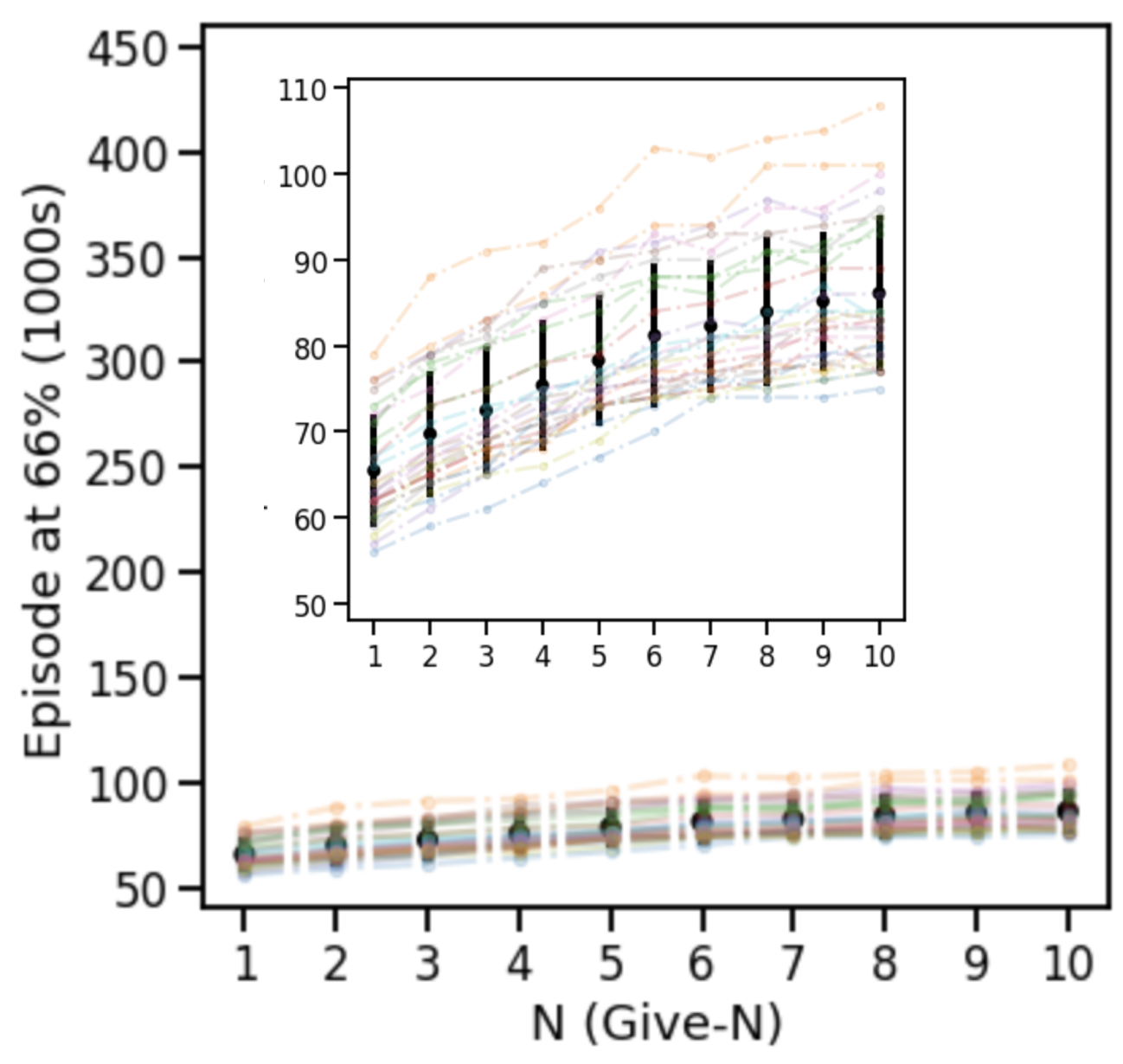}
  \subcaption{Dot-product Model}
  \label{dtraj}
\end{subfigure}
\begin{subfigure}{0.49\linewidth}
  \centering
  \includegraphics[width=\linewidth]{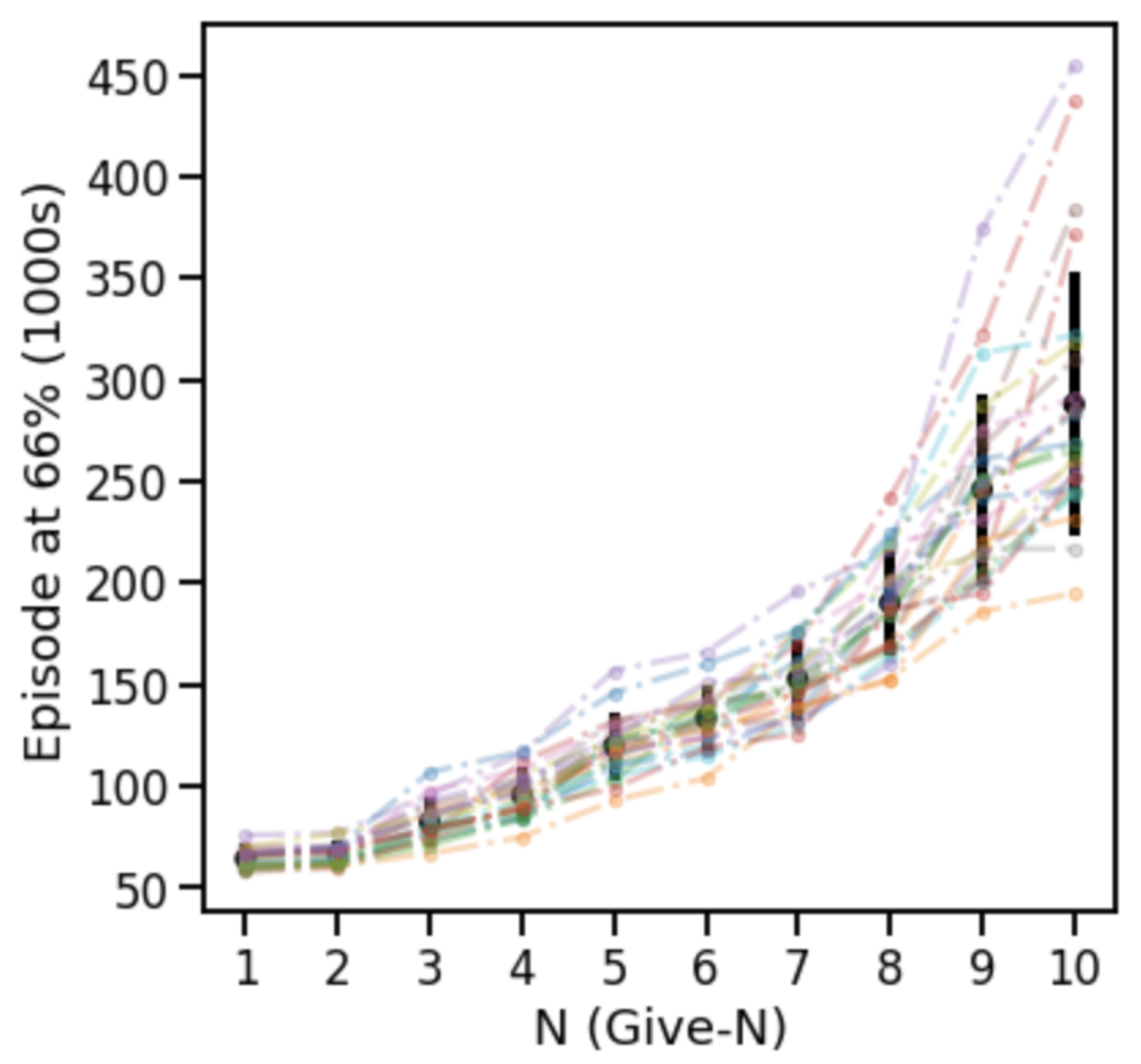}
  \subcaption{LSTM Baseline}
  \label{ltraj}
\end{subfigure}%
\begin{subfigure}{0.49\linewidth}
  \centering
  \includegraphics[width=\linewidth]{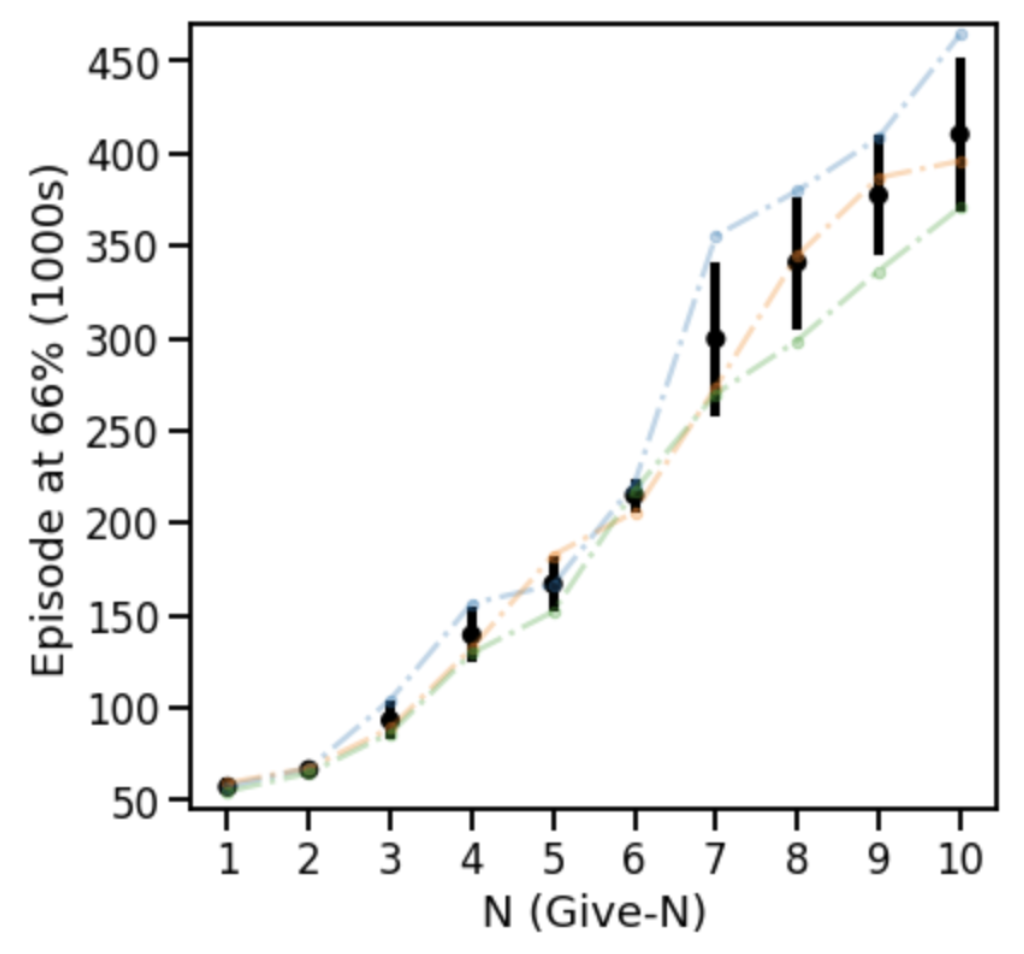}
  \subcaption{Transformer Baseline}
  \label{traj}
\end{subfigure}
\caption{Developmental trajectories for all models. For each requested $N$ on the x-axis, the episode at which a threshold of $66\%$ accuracy was crossed is displayed on the y-axis (which begins after 50,000 episodes of the step-1 curriculum). Insets show zoomed-in trajectories, colours represent individual models, and error bars represent standard deviation across models. Out of a training ensemble of 30 models, only those that reached threshold performance for all $N$ at some point during training were included (n=30/30 for ESBN and dot-product models, n=25/30 for LSTM baseline, and n=3/30 for the transformer baseline).}
\label{thresh}
\end{figure}




\section{Results}

Overall performance and extrapolation performance are shown in Figure \ref{extrap}. The ESBN model learned the task well, and achieved significant extrapolation. The dot-product model did even better in these respects. In contrast, while the baseline models performed well for $N$ up to around 5, performance degraded in various ways past that.  For example, unlike the ESBN and dot-product models, the LSTM baseline struggled to perform well for higher values of $N$ simultaneously, indicated by the difference between best accuracy at any time during training, and best accuracy when the model was doing its best on average. The transformer baseline struggled as $N$ increased, with only 3 models ever crossing threshold on Give-10. The baseline models were also incapable of extrapolation ($N$=11-15; Fig. \ref{bextrap} and \ref{textrap}).

The developmental trajectory of performance on the training set is shown in Figure \ref{thresh}.  In order to be included in the plot, a model had to have crossed threshold at some point during training for all $N$ displayed. The ESBN and dot-product models achieved criterial accuracy (66\%) for all values of $N$ for 30/30 models; in contrast, only 25/30 LSTM baseline and 3/30 transformer baseline models met this criterion. Though not shown, the transformer model also failed when we input the object vector after rather than before the transformer layer.

All models that met criterion displayed sequential learning, with higher values of $N$ crossing the threshold later than lower values of $N$ (this occurred even without being enforced by curriculum training, but those results are not shown). However, the ESBN and dot product models achieved these thresholds much earlier in training ($\sim 100k$ episodes) compared to the baseline models ($\sim 450k$ episodes).  However, only the ESBN displayed an inflection point, past which criterial performance for higher $N$'s were reached after many fewer epochs of training, and sometimes almost immediately. This was confirmed using a Bayesian information criterion to compare linear, exponential, logarithmic and sigmoidal fits to the development trajectory of the ESBN model. The sigmoidal fit best; and, when fit to the trajectory of individual instances of the model to quantify the $N$ of inflection, exhibited a mean of $4.38 \pm 0.39$ (mean $\pm$ sd). 




\section{Discussion}

We showed that a model of counting based on the ESBN architecture, and trained with reinforcement learning, exhibited a developmental trajectory qualitatively similar to the one observed in humans learning to count, as well as the capacity for systematic extrapolation. A model that implemented only the retrieval operation required for the counting task (the dot-product similarity operation) displayed good performance and extrapolation, but not a clear inflection in its developmental trajectory. Baseline models using either an LSTM or transformer as a controller, but without the external memory component, displayed much slower learning overall, no inflection in the learning trajectory, and no capacity for extrapolation. The transformer model did particularly poorly, possibly because standard transformers are ill-suited to processing adjacent time-steps \cite{mishra2017simple}.

One explanation for the success of the ESBN model was identified where it was first described \cite{webb2021emergent}.  There, the authors argued that because the information stream accessible to the controller was isolated from the incoming data stream by the key/value memory, the controller was free to produce and respond to abstract representations needed to perform the task, without being shaped or tied to individual items (tokens); these could be thought of as fulfilling the role of symbols in traditional architectures. Here, the key associated with (i.e. bound to) the instruction embedding $\bm{z_0}$ was this symbol, which functioned as $N$ in the give-$N$ task. Since the controller's job was to report when the correct count was reached, it simply had to recognize when $\bm{k_r}$ was close enough to this key. Once learned, it could quickly gain the capacity to give any $N$ for as high as it could count. 

Here, we relaxed the isolation of the controller from the data, giving it access to both the key and value streams at every time-point, and allowed it to learn which source of information was most useful. In principle, the network could have ignored the input from its external memory (the retrieved keys), performing the task only on the basis of the count embeddings that it received directly as input. However, this likely would have resulted in overfitting to the count embeddings observed in the training set, preventing extrapolation to new count embeddings, as was observed for the LSTM and transformer baseline models. Surprisingly, the ESBN did not display this overfitting, suggesting that it was indeed performing the task on the basis of information retrieved from its external memory. 

We hypothesize that this occurred because the gradients associated with the direct input of count embeddings at the beginning of an episode tended to vanish over the course of the episode, whereas the information retrieved from external memory was available at the time point immediately before the relevant action (\textit{done}) was taken. This suggests that the strict architectural separation in the original ESBN model might not be necessary to achieve systematic behavior. It is also interesting to note that both baseline models had good performance until around $N=5$, past which performance degraded, reflecting the possibility that increased task difficulty after this point pressured the transition from a specific to a systematic solution in the ESBN model. It could be that a similar mechanism is responsible for the transition seen around $N=5$ in children. 


We found that a simpler version of the model, incorporating only the dot-product similarity operation, displayed a comparable level of task performance and extrapolation. This suggests that, in the context of this task, this dot-product operation was the key inductive bias that enabled the ESBN to display systematic counting behavior. However, this simpler version of the model was specifically designed for the counting task, since the relevant similarity value (between the instruction and the count at each time step) was passed directly to the controller, rather than being embedded in a more general-purpose memory architecture. Furthermore, this version of the model did not exhibit an inflection in its learning trajectory, suggesting that this inflection may reflect the difficulty of learning to interact with memory. For these reasons, this model should be viewed as an attempt to better understand the operations of the ESBN, rather than a competing theoretical account.

One concern with the developmental trajectory displayed by the ESBN might be that, although it exhibits an inflection, that is softer than a more discrete transition to the CP-knower stage suggested by some developmental data \cite{wynn1992children}. It is worth emphasizing that this data was collected at longitudinal intervals of a minimum of 5-8 weeks, and so may not have adequate temporal resolution to clearly distinguish between these two possibilities. Future work involving more fine-grained longitudinal evaluation might test whether this transition is truly discrete, or closer to the sigmoidal trajectory displayed by the ESBN.

We exploited a form of curriculum learning in the present work, first pre-training networks to iterate through the count sequence, then training them to select objects, and finally to perform the give-$N$ task. It was possible for networks to learn all of these tasks at the same time, but we chose a curricular approach to mirror the fact that children are typically able to memorize arbitrary sequences early on, and often can count to numbers much higher than they are able to successfully employ in tasks such as give-$N$ \cite{fuson1982acquisition}. Importantly, the pre-learned count sequence could be used as scaffolding to support extrapolation to numbers outside the range of training on the give-$N$ task.

Some have argued that the transition to being a CP-knower, as measured by the give-$N$ task, marks a more general semantic induction of abstract number concepts, as measured by other related tasks (such as the ability to judge which of two numbers is greater) \cite{sarnecka2008counting}. Our model, which was only trained to perform the give-$N$ task, offers an alternative interpretation: the developmental trajectory observed in this task might reflect the learning of a systematic, but narrow, procedure, rather than a general understanding of cardinality. In line with this view, there is some evidence to suggest that children can often succeed on the give-$N$ task while failing to perform closely related tasks \cite{davidson2012does}. Thus, similar to our model, children may indeed undergo a phase in which their ability to perform related counting tasks is not yet integrated, and is better characterized as a set of ‘blind’ procedures specific to each task, despite being able to perform those procedures in a systematic manner. This is in line with the ‘knowledge-in-pieces’ view of development, whereby early concepts are not immediately integrated into a coherent whole \cite{disessa2014history}.

\subsubsection{Limitations and Future Work} 

Despite the limitation of having trained our models on only one task, it is possible that the ESBN architecture may not only facilitate the learning of systematic procedures in specific tasks (as demonstrated here) but in a multi-task learning context as well. In future work, we plan to train networks on multiple related tasks (e.g. the how-many task, unit task, and direction task \cite{sarnecka2008counting}) and study whether the ESBN affords a similar benefit in terms of the ability to perform these tasks systematically, and in a manner that mirrors the human developmental trajectory. We are also interested in allowing the controller to select the most useful stream of data to bind to its memory depending on its goal, so that we don't need to hard-code its use of the pre-trained counter. Preliminary data suggest the model is capable of making this selection.

Additionally, some instances of the baseline models failed to learn the full task, and some of the ESBN models were unstable, learning the task well initially but then having performance drop as training progressed.  This could have been due to the high-variance of the policy gradient estimator (the REINFORCE algorithm). We tried to replicate the human ability to learn from a single episode at a time, but off-policy learning methods (e.g. replay) might be required to ensure all networks reliably learn the task. As well, more advanced policy gradient algorithms might improve performance of our models, and are left to future work for implementation.

Finally, our model was restricted in that it could only count as high as the length of the count sequence it was trained to memorize, representing an intermediate level of human development (i.e. the stage when children cannot count higher than their memorized count sequence, despite being presumed CP-knowers). Most adults are capable of counting to arbitrarily large numbers, using a recursive, hierarchical algorithm; that is, cycling through digits and keeping track of place values. We believe that introducing hierarchical structure, context segmentation and normalization \cite{webb2020learning}, together with recursive application of the counter's ability to iterate through a fixed sequence, might permit the capacity for unbounded counting.  This offers the promise of providing a neurally plausible mechanism for symbolic counting, which lies at the heart of many powerful cognitive functions of which humans are capable, including mathematical reasoning.

\bibliographystyle{apacite}

\setlength{\bibleftmargin}{.125in}
\setlength{\bibindent}{-\bibleftmargin}

\bibliography{CogSci_Template}

\section{Acknowledgements}

This project / publication was made possible through the support of a grant from the John Templeton Foundation. Thank you to Steven Frankland, Simon Segert, Randall O’Reilly, and Alexander Petrov for their helpful discussions.

\end{document}